\documentclass[conference]{IEEEtran}
\IEEEoverridecommandlockouts

\usepackage{cite}
\usepackage{amsmath,amssymb,amsfonts}
\usepackage{algorithmic}
\usepackage{graphicx}
\usepackage{textcomp}
\usepackage{xcolor}

\usepackage{lipsum}
 
\newcommand\blfootnote[1]{%
  \begingroup
  \renewcommand\thefootnote{}\footnote{#1}%
  \addtocounter{footnote}{-1}%
  \endgroup
}

\def\BibTeX{{\rm B\kern-.05em{\sc i\kern-.025em b}\kern-.08em
    T\kern-.1667em\lower.7ex\hbox{E}\kern-.125emX}}
\begin{document}

\title{Sharpening Your Density Fields: Spiking Neuron Aided Fast Geometry Learning}

\author{
Yi Gu$^{*}$ \qquad  Zhaorui Wang$^{*}$ \qquad Dongjun Ye \qquad Renjing Xu$^\dag$ \\
HKUST(GZ)\\
{\tt\small \{ygu425, zwang408\}@connect.hkust-gz.edu.cn, renjingxu@hkust-gz.edu.cn}\\{\tt\small imath@omnispaceai.com}\\

}

\maketitle

\blfootnote{*  Equal Contribution}
\blfootnote{\dag\space  Corresponding author}

\begin{abstract}
Neural Radiance Fields (NeRF) have achieved remarkable progress in neural rendering. Extracting geometry from NeRF typically relies on the Marching Cubes algorithm, which uses a hand-crafted threshold to define the level set. However, this threshold-based approach requires laborious and scenario-specific tuning, limiting its practicality for real-world applications. In this work, we seek to enhance the efficiency of this method during the training time. To this end, we introduce a spiking neuron mechanism that dynamically adjusts the threshold, eliminating the need for manual selection. Despite its promise, directly training with the spiking neuron often results in model collapse and noisy outputs. To overcome these challenges, we propose a round-robin strategy that stabilizes the training process and enables the geometry network to achieve a sharper and more precise density distribution with minimal computational overhead. We validate our approach through extensive experiments on both synthetic and real-world datasets. The results show that our method significantly improves the performance of threshold-based techniques, offering a more robust and efficient solution for NeRF geometry extraction.
\end{abstract}


\section{Introduction}
\label{sec:intro}
Neural surface reconstruction from multi-view images is a fundamental challenge in computer vision and computer graphics. Neural Radiance Fields (NeRF)~\cite{mildenhall2020nerf} has achieved remarkable success in novel view synthesis, inspiring subsequent works~\cite{li2023neuralangelo, miller2024objects, wang2021neus, wu2022voxurf} to make significant advancements in neural surface reconstruction. To extract geometry from NeRF, most existing frameworks~\cite{peng2021neural, pumarola2021d, boss2021nerd} depend on the Marching Cubes~\cite{lorensen1998marching} algorithm, which requires a hand-crafted threshold to determine the level set. However, the optimal threshold often requires numerous attempts for different scenes. Moreover, this threshold-based approach can be less effective in low-density scenarios, where semi-transparent objects and thin structures are present.

To address this challenge, Spiking NeRF~\cite{liao2024spiking} applies a spiking neuron to dynamically adjust the density threshold during the training time. The main drawback of Spiking NeRF lies in the lengthy training process, restricting its application in practical scenarios. This raises a straightforward question: can the spiking neurons be applied to faster methods, i.e., grid-based neural fields?

\begin{figure}[ht]
    \centering
    \includegraphics[width = \linewidth]{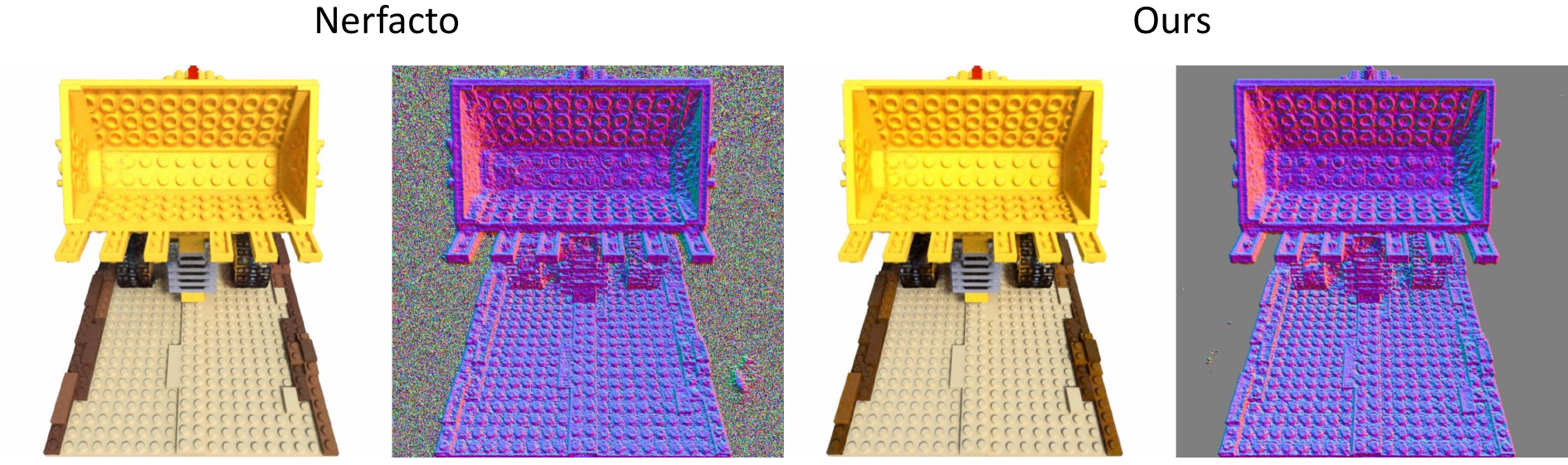}
    \caption{Visual comparisons of our method and Nerfacto~\cite{tancik2023nerfstudio}. Both models are trained for 10,000 iterations within 5 minutes, without the mask loss. As shown in the rendered color images and normal maps, our method extracts cleaner and more accurate geometry, while Nerfacto exhibits floating artifacts in the surrounding areas.
    }
    \label{fig:teaser}
\end{figure}

To investigate this problem, we directly apply the idea of the spiking neuron method (referred to as the spiking-threshold method) to Nerfacto~\cite{tancik2023nerfstudio}, a highly efficient and flexible framework. However, the experimental results reveal numerous pits and holes in reconstructed geometries, as shown in Fig.~\ref{fig:ablation}. Additionally, we empirically observe that this direct approach can lead to model collapse in certain cases, which involves a NaN (Not a Number) loss. 

This phenomenon prompts us to reconsider the spiking-threshold method, and our analysis identifies two main drawbacks. First, the spiking-threshold method can be viewed as a pruning strategy. The dynamically adjusted threshold encourages the network to retain the most important structures to fit the ground truth images. When the threshold becomes excessively large, it can lead to artifacts such as pits and holes. However, only surrogate gradients in the spiking neurons can prevent the threshold from becoming infinitely large, which is insufficient to constrain the threshold.  The second issue arises from the density network. In grid-based methods, the density network is typically implemented as a shallow multi-layer perceptron (MLP) for efficiency. Since density values below the threshold receive no gradients, the corresponding grid features cannot be updated. Nevertheless, the density MLP is optimized throughout the entire training process, meaning some "dead" grid features may occasionally contribute to the optimization. This instability in the training process is undesirable and can be catastrophic. 

Based on the above analysis, we propose our round-robin strategy to stabilize the training process. In each round, we first train the continuous MLPs with image-rendering and Eikonal losses to steer the main optimization direction. Subsequently, we train the same MLPs with the spiking neuron attached to the density output, while keeping the color network fixed. The rationale for fixing the color network is explained in Sec.~\ref{method:round-robin} and Sec.~\ref{ablation: round}. We validate our method on both synthetic and real-world datasets. The experimental results generally demonstrate that our training strategy can build a sharper density distribution for the geometry network with negligible computational cost. Fig.~\ref{fig:teaser} shows the superiority of our method.

Our contributions can be summarized as:

\begin{itemize}
    \item We introduce the spiking-threshold idea to the grid-based radiance fields to enable fast geometry learning.
    \item We propose a simple yet effective round-robin strategy to stabilize the training.
    \item Extensive experiments on both synthetic and real-world datasets demonstrate that our method can enhance the reconstruction performance of threshold-based methods with high fidelity and efficiency.
\end{itemize}

\section{Related Work}

\noindent\textbf{Neural Implicit Representations and Surface Reconstruction.} Neural implicit representations of 3D scenes gain attention for their continuity and high spatial resolution. Neural Radiance Fields (NeRF)~\cite{mildenhall2020nerf}, a pioneering approach for implicitly representing 3D scenes through neural networks, has demonstrated promising results in novel view synthesis. Nevertheless, its capability to reconstruct surfaces suffers from the absence of constraints on geometry~\cite{verbin2022ref}. To enhance the reconstruction quality, subsequent works (e.g., NeuS~\cite{wang2021neus} and IDR~\cite{yariv2020multiview}) depend on Signed Distance Functions (SDF) for better geometry representation. Our work focuses on the surface reconstruction of the NeRF-based models with a spiking-threshold method.

\noindent\textbf{Grid-based Neural Fields.} NeRF requires extensive computation time for both training and rendering. Grid-based methods have demonstrated state-of-the-art performance in both rendering quality and inference speed. For example, Instant-NGP~\cite{mueller2022instant}, DVGO~\cite{sun2022direct}, and FastNeRF~\cite{garbin2021fastnerf} apply advanced caching techniques to speed up the training of NeRF.  Plenoxels~\cite{fridovich2022plenoxels} and Relu-fields~\cite{karnewar2022relu} even do not include any neural network and directly optimize the explicit 3D models. Moreover, grid-based approaches are not susceptible to the spectral bias inherent in neural networks, often resulting in better performance compared to MLP-based methods~\cite{rahaman2019spectral, nam2023mip}. Nerfacto~\cite{tancik2023nerfstudio} is proposed as a mix of methods taken from various publications. Our work is based on Nerfacto and focuses on extracting geometry from grid-based neural fields.

\noindent\textbf{Spiking Neural Networks (SNNs) in NeRF.} Recent studies highlight the great potential of bio-inspired NeRF~\cite{passos2024bionerf}. Some efforts~\cite{yao2023spiking} have integrated SNNs into the NeRF architecture, achieving high-quality results with low energy consumption. In the domain of surface reconstruction, Spiking NeRF~\cite{liao2024spiking} introduces spiking neurons into NeRF to adaptively update the density threshold, improving geometric quality. Similarly, Spiking GS~\cite{zhang2024spiking} proposes to utilize global and local spiking neurons to handle low-opacity Gaussians and the low-opacity tails of Gaussians. Our work extends this spiking-threshold concept to grid-based representations with a stable round-robin training strategy.

\section{Method}

\begin{figure*}[ht]
    \centering
    \includegraphics[width=0.98\linewidth]{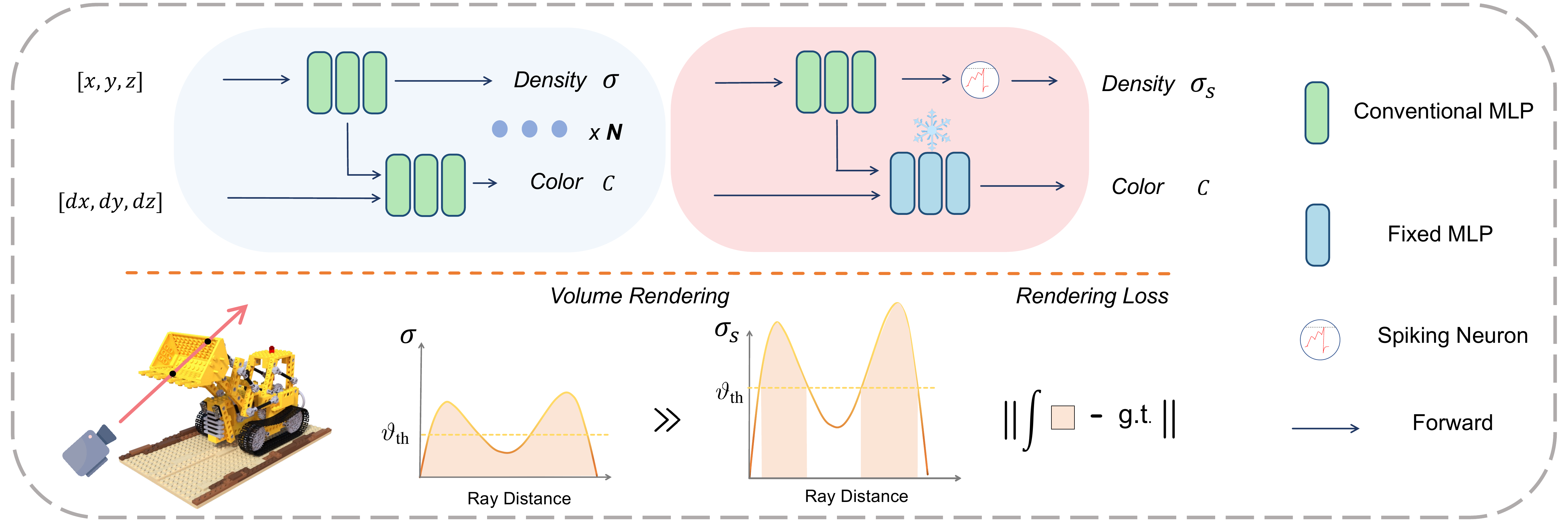}
    \caption{Framework overview of our method. The top of this figure is the pipeline, which depicts the network architecture with the round-robin strategy. The blue region represents the normal stage and the red region is the spiking stage. The bottom of the figure illustrates the loss computation process. The orange region means density values participate in volume rendering.}
    \label{fig:fig_1}
\end{figure*}

\subsection{Preliminary}

\noindent\textbf{Grid-based Radiance Fields.} For each 3D position $\mathbf{x}\in \mathbb{R}^3$, we use multi-resolution hash encodings~\cite{mueller2022instant} $h_\Omega(\mathbf{x})$ to learn the high-frequency spatial information, with learnable hash table entries $\Omega$. Then, our density network can be represented as:
\begin{equation} \label{eqn:density_net}
(\sigma,\ \mathbf{g}) = f_{\theta}(h_\Omega(\mathbf{x})),
\end{equation}

where $f_{\theta}$ is a shallow MLP with weights $\theta$, $\mathbf{g}$ is the geometry feature, and $\sigma$ represents the density value.

Given the density network, the normal of $\mathbf{x}$ can be computed as:
\begin{equation} \label{eqn:normal}
\mathbf{n}=-\frac{\nabla_{\mathbf{x}}\sigma}{||\nabla_{\mathbf{x}}\sigma||},
\end{equation}
where $\nabla_{\mathbf{x}}\sigma$ denotes the gradient of the density with respect to $\mathbf{x}$. As we focus on geometry extraction, this normal can be utilized to regularize our density network, as will be introduced in the loss function parts.

The color network is defined as:
\begin{equation} \label{eqn:color_net}
\mathbf{c} = c_{\Upsilon}(\mathcal{SH}(\textbf{d}),\ \mathbf{g}),
\end{equation}
where $c_{\Upsilon}$ is a shallow MLP with weights $\Upsilon$. $\mathcal{SH}(\textbf{d})$ is the spherical harmonics basis of the viewing direction $\textbf{d}$, which is a natural frequency encoding over unit vectors. By utilizing the volume rendering equation~\cite{mildenhall2020nerf}, we can construct the radiance fields efficiently with the supervision of the ground truth images.

\noindent\textbf{Spiking Neuron.}To get full-precision information, we employ the full-precision integrate-and-fire (FIF) spiking neuron~\cite{li2022brain, liao2024spiking, zhang2024spiking} in this work, which is widely used in the regression tasks. The definition of FIF neuron can be described as:  
\begin{equation}
    o = \Theta (u - \vartheta_{\textrm{th}}) * u
    = \begin{cases}
    0& \text{$u < \vartheta_{\textrm{th}}$},\\
    u& \text{otherwise},
    \end{cases}
     \label{eq:fif}
\end{equation}
where $u$ is the input value; $o$ is the output value and $\Theta(\cdot)$ is the Heaviside function. When the input value exceeds the threshold $\vartheta_{\textrm{th}}$, the output $o$ will equal the input $u$. 

\subsection{Our method}

\noindent\textbf{Spiking Neuron in Our Method.} To eliminate the tedious manual selection of the density threshold in NeRF geometry extraction, we propose to introduce the FIF neuron to dynamically adjust the density threshold.

The spiking neuron is attached after the density output $\sigma$, serving as a pruning strategy for density values. We annotate the original density value as $\sigma$ and the density filtered by the spiking neuron as $\sigma_{s}$. By employing the Eq.~\ref{eq:fif}, the truncated density value $\sigma_{s}$ can be calculated as:
\begin{equation}
    \label{eq:spiking_density}
    \sigma_{s}=\Theta(\sigma - \vartheta_{\textrm{th}}) * \sigma,
\end{equation}
where $\vartheta_{\textrm{th}}$ is the learnable threshold parameter. We initialize $\vartheta_{\textrm{th}}$ as $0$ to facilitate optimization during the early stage of training. To tackle the non-differentiable problem of the Heaviside function $\Theta$, we employ the surrogate functions to compute gradients. Thus, the gradients of the threshold and the input can be approximated by:
\begin{align}
\frac{\partial \sigma_{s}}{\partial \vartheta_{\textrm{th}}} = - r * \max(0, \frac{k-\left|\sigma - \vartheta_{\textrm{th}}\right|}{k^2}) * \sigma,\label{eq:diff_vth}
\end{align}
\begin{align}
    \frac{\partial \sigma_{s}}{\partial \sigma} 
    & = r * \max(0, \frac{k - |\sigma - \vartheta_{\textrm{th}}|}{k^2})
    * \sigma + \Theta(\sigma - \vartheta_{\textrm{th}}) \label{eq:diff_input},
\end{align}
where $k$ and $r$ are hyperparameters. It is necessary to incorporate Eq.\ref{eq:diff_vth} as the surrogate gradients, which helps regularize the threshold and prevent it from becoming excessively large.

We empirically find that the first term in Eq.~\ref{eq:diff_input} (i.e., $r * \max(0, (k - |\sigma - \vartheta_{\textrm{th}}|)/{k^2}) * \sigma$) often results in ambiguous outcomes, where the colors are accurate but the geometries are incorrect. Therefore, we only keep the second term for backpropagation (i.e., $\partial \sigma_{s}/\partial \sigma=\Theta(\sigma - \vartheta_{\textrm{th}})$).

\noindent\textbf{Round-robin Strategy.} 
\label{method:round-robin}
As analyzed in Sec.~\ref{sec:intro}, the under-constraint dynamic threshold may result in unstable training. Drawing from the trial-and-error approach, we propose a simple yet effective round-robin strategy to address this issue. As illustrated in the pipeline in Fig.~\ref{fig:fig_1}, we define a round as a sequential of $N+1$ iterations. Each round is divided into two stages, i.e., the normal stage and the spiking stage. In the normal stage, the full range of density values is used to construct the radiance fields, while the spiking stage operates on the truncated density values. The spiking stage aims to exclude density values lower than the learnable threshold $\vartheta_{\textrm{th}}$. This operation helps eliminate floaters and better explore the underlying geometries but may introduce undesired artifacts or lead to model collapse. The normal stage acts as a remedy to ensure the output of the density network remains reasonable and guides the main direction of the optimization. As training progresses, the density distribution gets sharpened around the surface, as illustrated at the bottom of Fig.~\ref{fig:fig_1}.

It is important to note that the color network in the spiking stage is fixed to avoid conflicts during the optimization. We validate the effectiveness of this fixed color operation in the ablation studies (Sec.~\ref{ablation: fix}).

\begin{figure*}[ht]
    \centering
    \includegraphics[width = \linewidth]{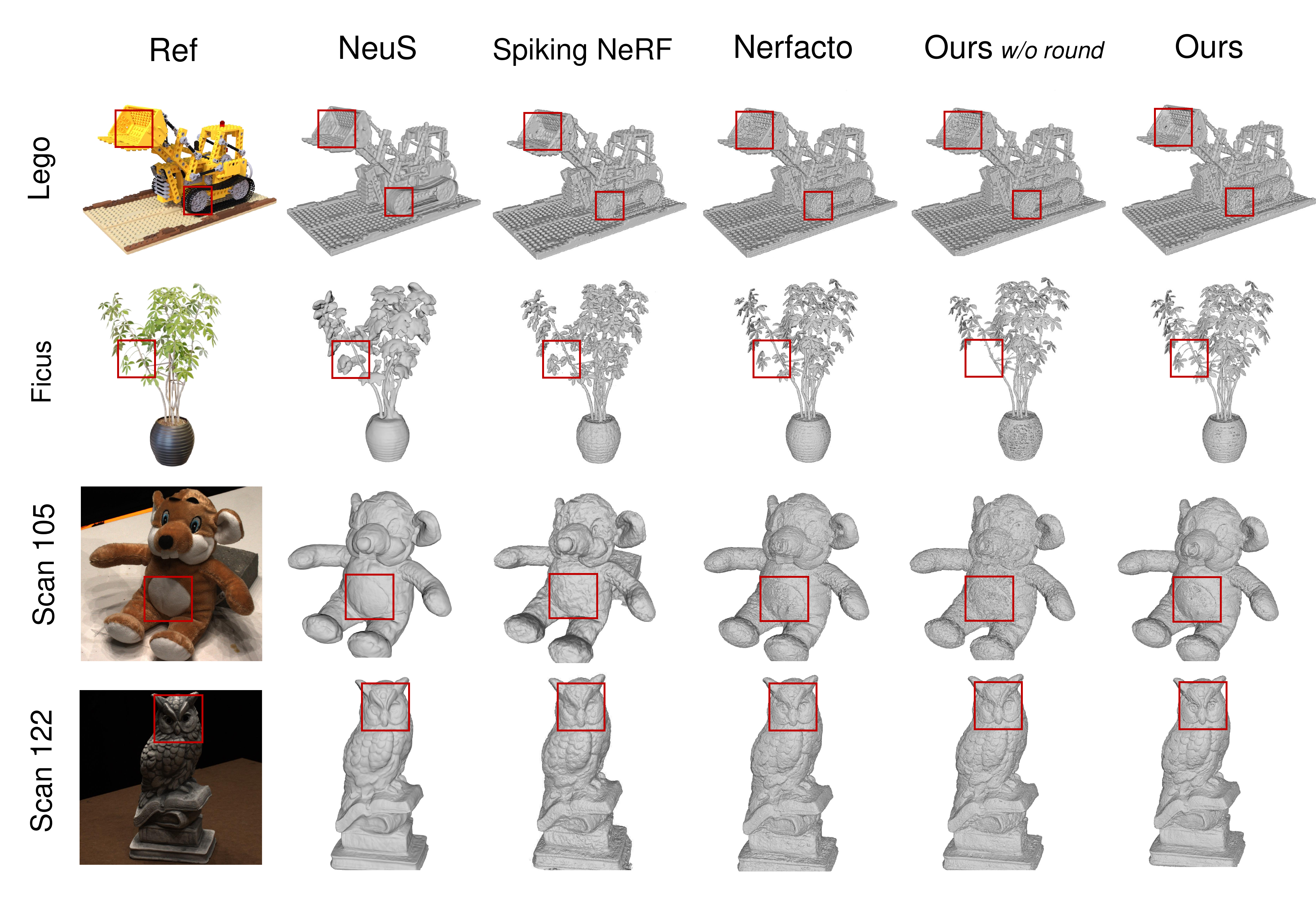}
    \caption{Visual Quality Comparisons on the Blender dataset~\cite{mildenhall2020nerf} and the DTU dataset~\cite{jensen2014large}. Red bounding boxes are used to highlight areas with significant differences. Our method can reconstruct more high-quality geometries. The full set of visual comparisons is provided in the supplementary materials.}
    \label{fig:fig_2}
\end{figure*}

\noindent\textbf{Loss Functions.} Our overall loss $\mathcal{L}$ is defined as follow:
\begin{align}
    \mathcal{L} = \mathcal{L}_{c} + \mathcal{L}_v + \mathcal{L}_{optional}.
\end{align}

The color loss $\mathcal{L}_{c}$ is defined as:
\begin{align}
    \mathcal{L}_{c} = \frac{1}{m}\sum_{\mathbf{r}\in \mathcal{R}} ||\hat{C}(\mathbf{r}) - C(\mathbf{r})||,
\end{align}
with $\mathcal{R}$ representing the set of $m$ rays in each training batch and $\mathbf{r}$ representing the sampled ray. $\hat{C}$ is the rendered color, which can be obtained by the volume rendering equation~\cite{mildenhall2020nerf}, and $C$ is the ground truth color.

The $\mathcal{L}_v$ loss is the threshold loss, which is proposed to enhance the effect of FIF neurons by encouraging the density threshold to increase. It is defined as follows:
\begin{align}
    \mathcal{L}_v = \frac{\lambda_{v}}{e^{\vartheta_{\textrm{th}}}},
\end{align}
where $\lambda_{v}$ is a hyperparameter.

We also provide three optional losses to achieve better geometry results. Following Ref-NeRF~\cite{verbin2022ref} ans NeuS~\cite{wang2021neus}, incorporating the orientation loss and Eikonal loss~\cite{gropp2020implicit} can improve the smoothness of the geometry network. Furthermore, we can also utilize mask supervision for cleaner geometric results. Thus, our optional loss $\mathcal{L}_{optional}$ can be defined as:

\begin{align}
    \mathcal{L}_{optional} =\lambda_{o} \mathcal{L}_{o} + \lambda_{eik}\mathcal{L}_{eik} + \lambda_{m}\mathcal{L}_m,
\end{align}
where $\lambda_o$, $\lambda_{eik}$ and $\lambda_m$ are hyperparameters for the orientation loss $\mathcal{L}_{o}$, Eikonal loss $\mathcal{L}_{eik}$ and mask loss $\mathcal{L}_{m}$, respectively.

The orientation loss~\cite{verbin2022ref} is proposed to encourage that all visible normals are facing toward the camera, defined as:
\begin{align}
    \mathcal{L}_{o} = \sum_{i}w_i \max(0, \mathbf{n}_i \cdot \mathbf{d})^2,
\end{align}
where $i$ is the index of sampled points along a ray with the view direction $\mathbf{d}$. $\mathbf{n}_i$ is the normal vector and $w_i$ is the weight of the $i$-th point.

The Eikonal loss is proposed in~\cite{gropp2020implicit}, defined as:
\begin{align}
    \mathcal{L}_{eik} = \frac{1}{N} \sum_{i}(||\nabla_{\mathbf{x_i}}\sigma||_2 -1)^2,
\end{align}
where $N$ is the number of sampled points and $i$ is the index of points.

If the mask is available, we can also introduce the mask loss $\mathcal{L}_m$ to better extract interest objects. We use binary cross-entropy ($BCE$) to compute $\mathcal{L}_m$:
\begin{align}
    \mathcal{L}_{m} = BCE(M_k, \hat{O}_k),
\end{align}
where $M_k$ is the ground truth mask of the $k$-th view and $\hat{O}_k$ is the accumulation weight.

\begin{table*}[htbp]
\centering
\setlength{\tabcolsep}{7pt} 
\renewcommand{\arraystretch}{1.4}
\caption{Quantitative Comparisons on the DTU dataset~\cite{jensen2014large}. We measure the performance by Chamfer distance. The \textbf{\textcolor{red}{red}} and \textbf{\textcolor{blue}{blue}} numbers indicate the first and the second performer for each scene.}
\begin{tabular}{lcccccccccccccccc}
\hline
\cline{1-17} 
       \textbf{Scan} & \textbf{24} & \textbf{37} & \textbf{40} & \textbf{55} & \textbf{63} & \textbf{65} & \textbf{69} & \textbf{83} & \textbf{97} & \textbf{105} & \textbf{106}& \textbf{110} & \textbf{114} & \textbf{118} & \textbf{122} & \textbf{Avg.}\\ 
\hline
NeuS~\cite{wang2021neus}     & 1.00 & 1.37 & \textbf{\textcolor{blue}{0.93}}  & \textbf{\textcolor{blue}{0.43}} & 1.10 & \textbf{\textcolor{red}{0.70}} & \textbf{\textcolor{red}{0.57}} & 1.48 & \textbf{\textcolor{red}{1.16}} & \textbf{\textcolor{blue}{0.83}} & \textbf{\textcolor{blue}{0.52}} & 1.69 & \textbf{\textcolor{red}{0.35}}  & \textbf{\textcolor{red}{0.49}} & \textbf{\textcolor{blue}{0.54}} & \textbf{\textcolor{blue}{0.88}}\\
SpNeRF~\cite{liao2024spiking}     &  \textbf{\textcolor{red}{0.84}} & \textbf{\textcolor{red}{1.20}} & 1.02 & \textbf{\textcolor{red}{0.38}} & 1.15 & \textbf{\textcolor{blue}{0.72}} & \textbf{\textcolor{blue}{0.69}} & \textbf{\textcolor{red}{1.10}} & \textbf{\textcolor{blue}{1.19}} & \textbf{\textcolor{red}{0.65}} & \textbf{\textcolor{red}{0.49}} & 1.60 & \textbf{\textcolor{blue}{0.49}} & \textbf{\textcolor{blue}{0.55}} &\textbf{\textcolor{red}{0.51}} & 
\textbf{\textcolor{red}{0.83}}\\
Nerfacto~\cite{tancik2023nerfstudio}  & 0.98  & 1.44 & 0.95 & 0.49 & \textbf{\textcolor{blue}{1.06}} & 1.06 & 1.08 & 1.26 & 1.49 & 1.42 & 0.88 & \textbf{\textcolor{blue}{1.36}} & 1.08 & 0.82 & 0.72 & 1.01\\
Ours     & \textbf{\textcolor{blue}{0.94}} & \textbf{\textcolor{blue}{1.25}} & \textbf{\textcolor{red}{0.81}} & 0.48 & \textbf{\textcolor{red}{1.00}} & 0.90 & 1.14 & \textbf{\textcolor{blue}{1.21}} & 1.70 & 1.17 & 0.73 & \textbf{\textcolor{red}{1.33}} & 1.14 &  0.85 & 0.53 & 0.94\\

\hline
\cline{1-17} 
\end{tabular}
\label{table:dtu}
\end{table*}

\begin{table}[htbp]
\centering
\scriptsize
\setlength{\tabcolsep}{3.4pt} 
\renewcommand{\arraystretch}{1.5}
\caption{Quantitative Comparisons on the Blender dataset~\cite{mildenhall2020nerf}. We measure the performance by Chamfer distance $\times 10^{-2}$.}
\begin{tabular}{lccccccccc}
\hline
\cline{1-10} 
       \textbf{} & \textbf{Lego} & \textbf{Chair} & \textbf{Mic} & \textbf{Ficus} & \textbf{Hotdog} & \textbf{Drums} & \textbf{Mats} & \textbf{Ship} & \textbf{Avg.} \\ 
\hline
NeuS~\cite{wang2021neus}     & 1.52 & 0.70 & 0.85 & 1.67 & 1.40 & 4.27 & 1.08 & 2.33 & 1.73 \\
SpNeRF~\cite{liao2024spiking}     & \textbf{\textcolor{blue}{0.70}} & \textbf{\textcolor{blue}{0.66}} & 0.72 & 0.54 & 0.94 & 2.43 & 1.10 & \textbf{\textcolor{blue}{1.49}} & 1.07 \\
Nerfacto~\cite{tancik2023nerfstudio}   & 0.77 & \textbf{\textcolor{red}{0.58}} & \textbf{\textcolor{blue}{0.41}} & \textbf{\textcolor{blue}{0.49}} & \textbf{\textcolor{blue}{0.79}} & \textbf{\textcolor{blue}{0.81}} & \textbf{\textcolor{blue}{0.62}} & 1.66 & \textbf{\textcolor{blue}{0.77}} \\

Ours    & \textbf{\textcolor{red}{0.64}} & 0.75 & \textbf{\textcolor{red}{0.40}} & \textbf{\textcolor{red}{0.43}} & \textbf{\textcolor{red}{0.69}} & \textbf{\textcolor{red}{0.63}} & \textbf{\textcolor{red}{0.41}} & \textbf{\textcolor{red}{1.25}} & \textbf{\textcolor{red}{0.65}} \\
\hline
\cline{1-10} 
\end{tabular}
\label{table:blender}
\end{table}

\begin{table}[htbp]
\centering
\scriptsize 
\setlength{\tabcolsep}{4.6pt} 
\renewcommand{\arraystretch}{1.2}
\caption{Ablation studies on  the Blender~\cite{mildenhall2020nerf} and the DTU~\cite{jensen2014large} datasets.}
\begin{tabular}{lccc|lccc}
\hline
\cline{1-8}
\multicolumn{4}{c|}{\textbf{Blender}~\cite{mildenhall2020nerf}} & \multicolumn{3}{c}{\textbf{DTU}~\cite{jensen2014large}} \\

\textbf{Scene} & \textbf{w/o round} & \textbf{w/o fix} & \textbf{full} & \textbf{Scene} & \textbf{w/o round} & \textbf{w/o fix} & \textbf{full}
       \\ 
\hline
\textbf{Lego}   & 0.81 & 0.69 & 0.64  & \textbf{scan 40} & 0.96 & 1.04 & 0.81 \\ 
\textbf{Drums}  & 0.76 & 1.11 & 0.63 & \textbf{scan 65} & 1.05 & 0.93 & 0.91\\
\textbf{Ficus}  & 1.03 & 0.47 & 0.43 & \textbf{scan 106} & 0.80 & 0.85 & 0.74 \\
\textbf{Hotdog} & 1.58 & 1.21 & 0.69 & \textbf{scan 122} & 0.57 & 0.72 & 0.53 \\

\hline
\cline{1-8} 
\end{tabular}
\label{table:ablation}
\end{table}

\section{Experiments}
\subsection{Experimental Settings}
\noindent\textbf{Datasets, Metrics and Baselines.} We validate our approach on the Blender dataset~\cite{mildenhall2020nerf} and the DTU dataset~\cite{jensen2014large}. The Blender dataset contains 8 synthetic scenes with various lighting conditions. The DTU dataset contains 15 real-world scenes, captured from limited and fixed views. We compare our method with 3 baselines, i.e., NeuS~\cite{wang2021neus}, Spiking NeRF~\cite{liao2024spiking}, and Nerfacto~\cite{tancik2023nerfstudio}. Following the literature of neural surface reconstruction~\cite{wang2021neus, liao2024spiking}, we use the Chamfer distance (CD) as the main metric. 

\noindent\textbf{Implementation Details.}
We implement our method based on the Nerfacto~\cite{tancik2023nerfstudio} framework. We set $\lambda_v = 0.05$, $\lambda_o = 0.0001$, $\lambda_{eik} = 0.0001$ and $\lambda_m = 1$ for all experiments. Each scene is trained for 30k iterations, the same as the Nerfacto baseline. All experiments are conducted on a single NVIDIA RTX3090 GPU. Our method runs an average of 15 minutes in the Blender dataset and 10 minutes in the DTU dataset. At the end of training, we directly employ the Marching Cubes algorithm~\cite{lorensen1998marching} to extract the mesh surface, with the level set defined as $\vartheta_{\textrm{th}}$.

\subsection{Quantitative Comparisons}
The quantitative results of the DTU dataset~\cite{jensen2014large} are presented in Tab.~\ref{table:dtu}. While NeuS~\cite{wang2021neus} and Spiking NeRF~\cite{liao2024spiking} achieve higher overall performance compared to our method and Nerfacto~\cite{tancik2023nerfstudio}, they require significantly longer training time — exceeding 10 hours on the same device. Another key difference lies in the background modeling. Unlike our approach, NeuS and Spiking NeRF follow the NeRF++~\cite{zhang2020nerf++} method, constructing an additional radiance field to model the background, which is beyond the scope of our method. Despite these differences, our method achieves the best performance on three scenes and the second-best performance on another three. Moreover, our method consistently outperforms Nerfacto, demonstrating its effectiveness.

The quantitative results of the Blender dataset~\cite{jensen2014large} are reported in Tab.~\ref{table:blender}. As the Blender dataset is captured in 360 degree with a pure background, background modeling has less impact on the final geometry. As a result, NeuS~\cite{wang2021neus} and Spiking NeRF~\cite{liao2024spiking} show less competitive performance on this dataset. Our method surpasses all competitors, achieving the best performance on seven out of eight scenes. Although Nerfacto also delivers comparable results, we encourage readers to refer to the qualitative results for a more comprehensive comparison.

\begin{figure}[ht]
    \centering
    \includegraphics[width = \linewidth]{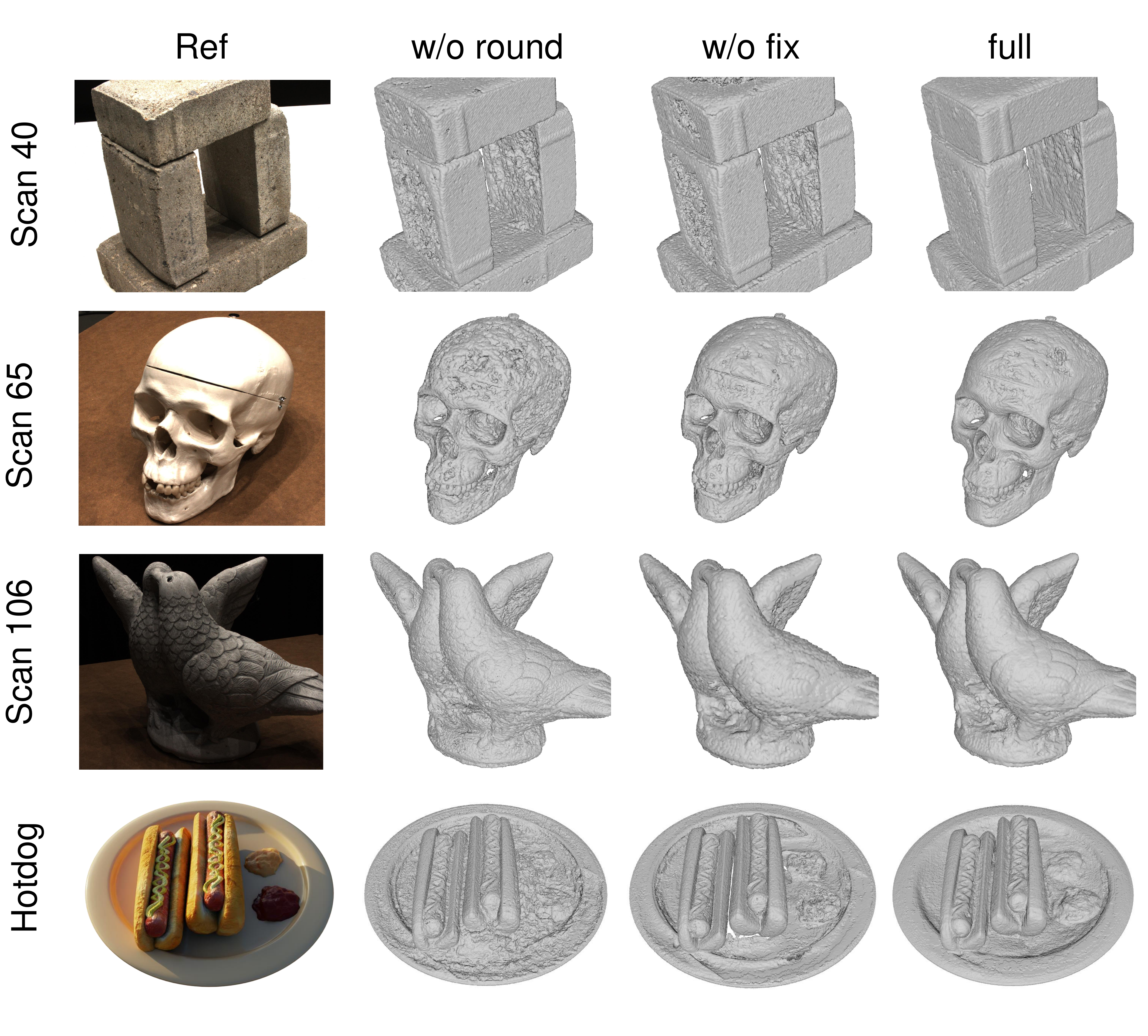}
    \caption{Ablation studies on the DTU dataset~\cite{jensen2014large} and the Blender dataset~\cite{mildenhall2020nerf}.}
    \label{fig:ablation}
\end{figure}

\subsection{Qualitative Comparisons}
We present sampled qualitative results in Fig.~\ref{fig:fig_2}, with the full set of visual results available in the supplementary materials. NeuS~\cite{wang2021neus} generally produces smooth results on both synthetic and real-world datasets but struggles to capture high-frequency details. Benefiting from the discontinuous representations, Spiking NeRF~\cite{liao2024spiking} and Nerfacto~\cite{tancik2023nerfstudio} reveal more details than NeuS but may miss certain parts, such as the branches in Ficus and the wheels in Lego. Directly applying spiking neurons without the round-robin strategy results in pits and holes, as observed in the 5th column of Fig.~\ref{fig:fig_2}. Compared to baseline methods, our approach achieves a more balanced performance, offering better completeness and smoothness in the reconstructed geometry.

\subsection{Ablation Study}
We select 4 cases from the Blender dataset~\cite{mildenhall2020nerf} and 4 cases from the DTU dataset~\cite{jensen2014large} to validate the effectiveness of each component. The visual comparisons are presented in Fig.~\ref{fig:ablation} and the quantitative results are presented in Tab.~\ref{table:ablation}. `\textit{w/o} round' and `\textit{w/o} fix' represent our method without the round-robin strategy and without fixing the color network respectively.

\noindent{}\textbf{Effectiveness of Round-robin Strategy.}\label{ablation: round} The visual results in the second column of Fig.~\ref{fig:ablation} reveal that surfaces reconstructed using only the spiking-threshold method exhibit significant noise. In contrast, our full model significantly enhances the geometry quality, both visually and in terms of the quantitative Chamfer distances. These improvements demonstrate the effectiveness of the round-robin strategy.

\noindent{}\textbf{Effectiveness of Fixed Color Network.}\label{ablation: fix}
The visual results in the third column of Fig.~\ref{fig:ablation} show that the `\textit{w/o} fix' configuration produces noticeable pits and hole artifacts. For example, the plate in the Hotdog scene appears ambiguous, while our full model achieves more accurate and smoother results. These observations highlight the importance of fixing the color network to ensure stable and consistent geometry reconstruction.

\section{Conclusion and Limitation}
In this paper, we propose a novel method for surface reconstruction from grid-based radiance fields. To eliminate the need for manual threshold selection, we introduce a spiking neuron mechanism that dynamically adjusts the threshold during training. Additionally, we design a simple yet effective round-robin strategy to stabilize the training process. Comprehensive experiments demonstrate that our approach efficiently extracts precise geometry from the grid-based NeRF pipeline.

Despite its effectiveness, our approach has certain limitations. One challenge is its difficulty in handling scenes with high light intensity or low brightness in real-world scenarios. A potential solution is to incorporate more advanced view-dependent parameterization methods~\cite{verbin2022ref} into our framework. Another limitation lies in our implementation of the Eikonal loss using the PyTorch framework, which slightly slows down training speed. Future work could address this by fully integrating the implementation with tiny-cuda-nn~\cite{wang2023neus2, mueller2022instant} to enable faster training.

\bibliographystyle{IEEEbib}
\bibliography{icme2025references}

\section{Supplemantary Materials}
\subsection{More Visual Results}

\begin{figure}[ht]
    \centering
    \includegraphics[width = \linewidth]{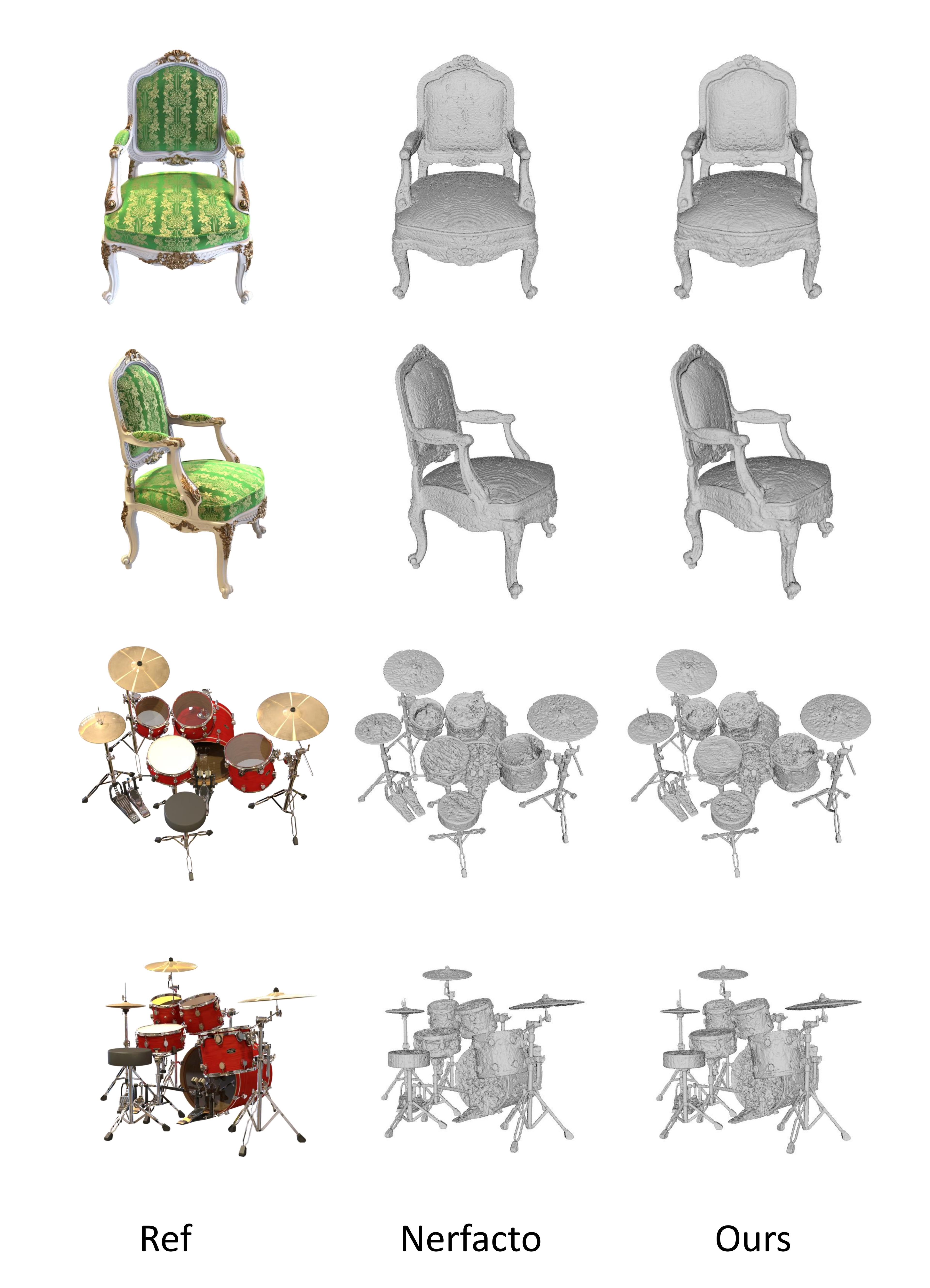}
    \caption{Reconstruction results on the Blender dataset~\cite{mildenhall2020nerf} (Chair, Drums).}
    \label{fig:supp_1}
\end{figure}

\begin{figure}[ht]
    \centering
    \includegraphics[width = \linewidth]{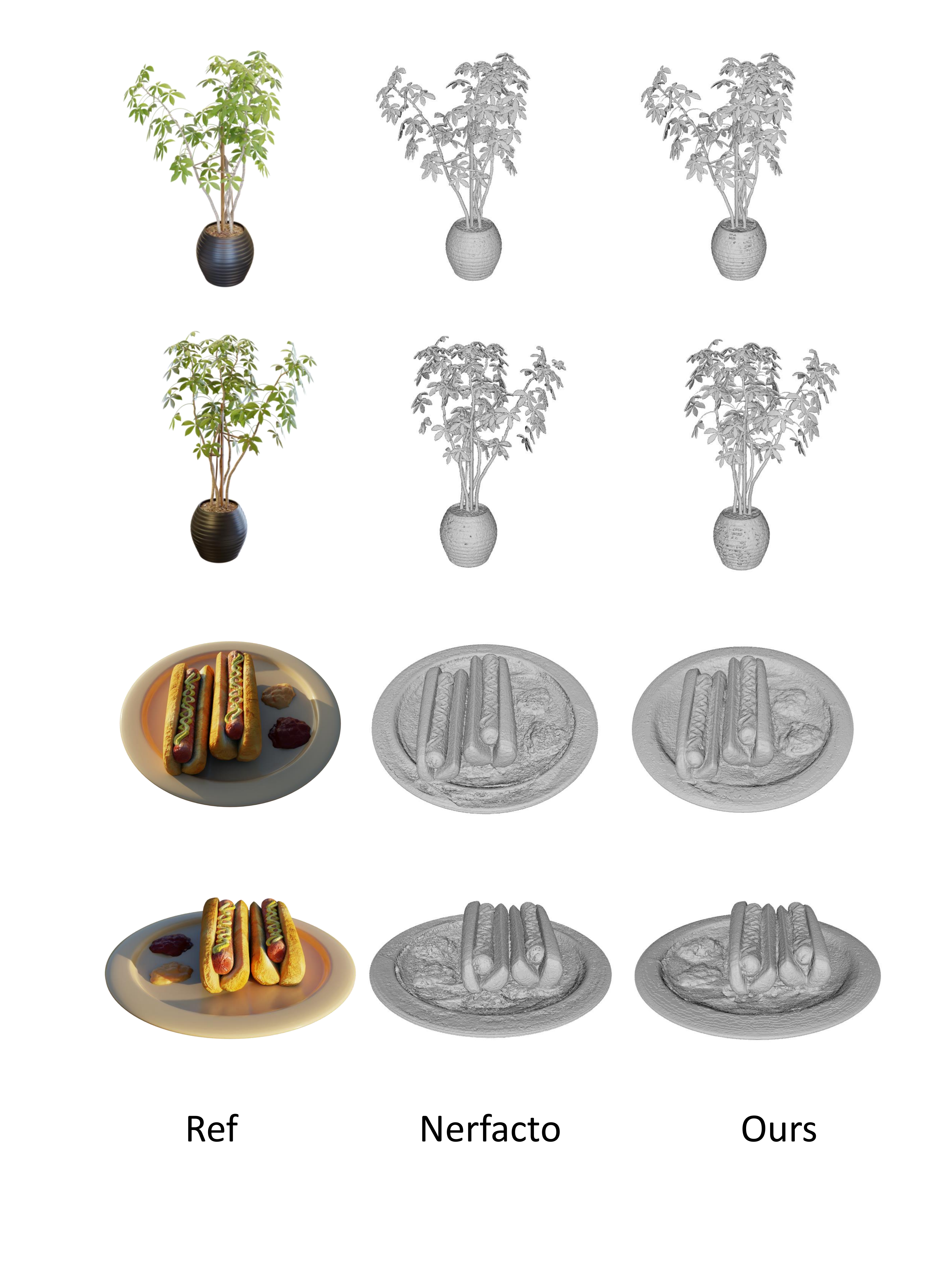}
    \caption{Reconstruction results on the Blender dataset~\cite{mildenhall2020nerf} (Ficus, Hotdog).}
    \label{fig:supp_2}
\end{figure}

\begin{figure}[ht]
    \centering
    \includegraphics[width = \linewidth]{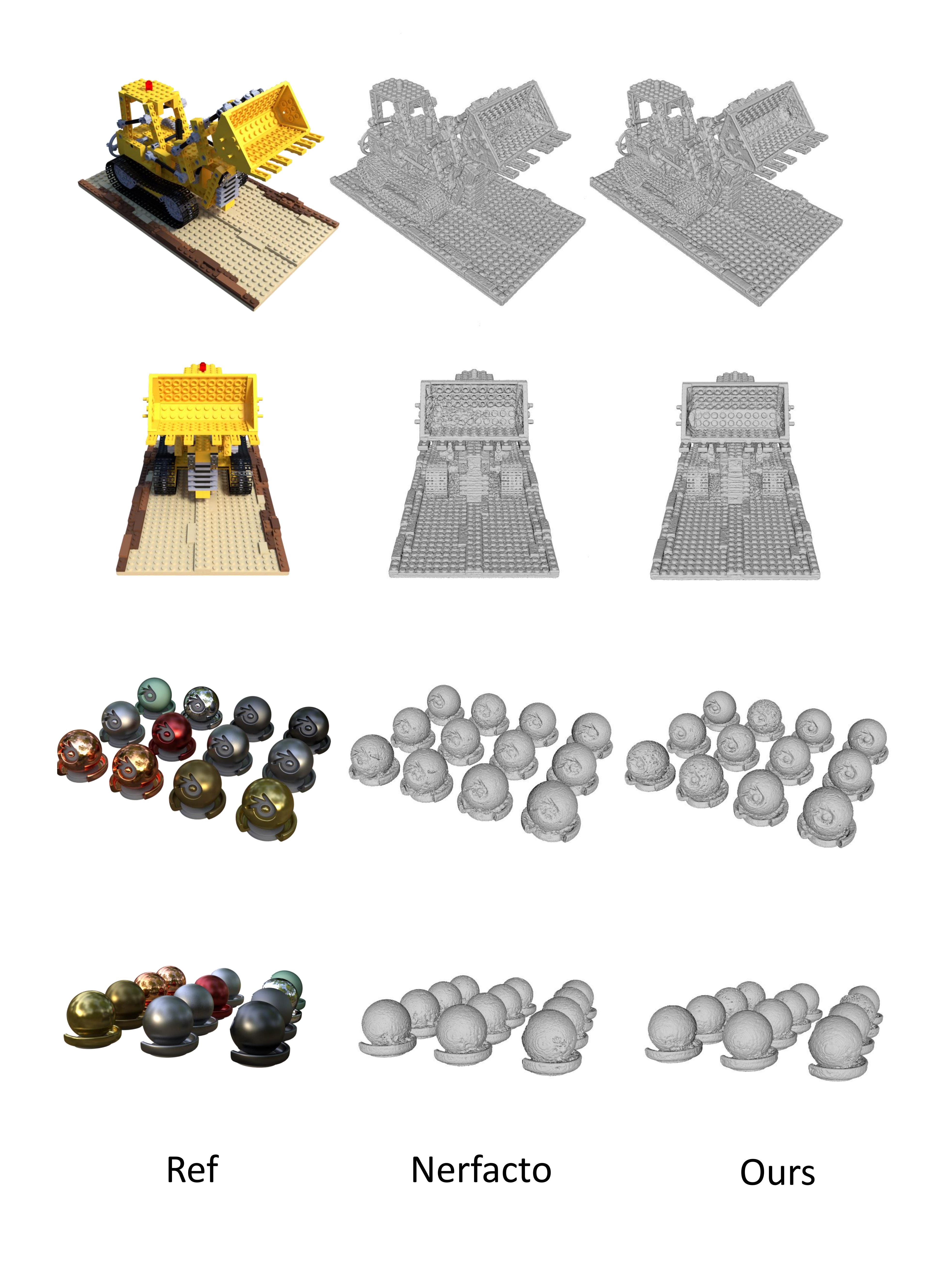}
    \caption{Reconstruction results on the Blender dataset~\cite{mildenhall2020nerf} (Lego, Materials).}
    \label{fig:supp_3}
\end{figure}

\begin{figure}[ht]
    \centering
    \includegraphics[width = \linewidth]{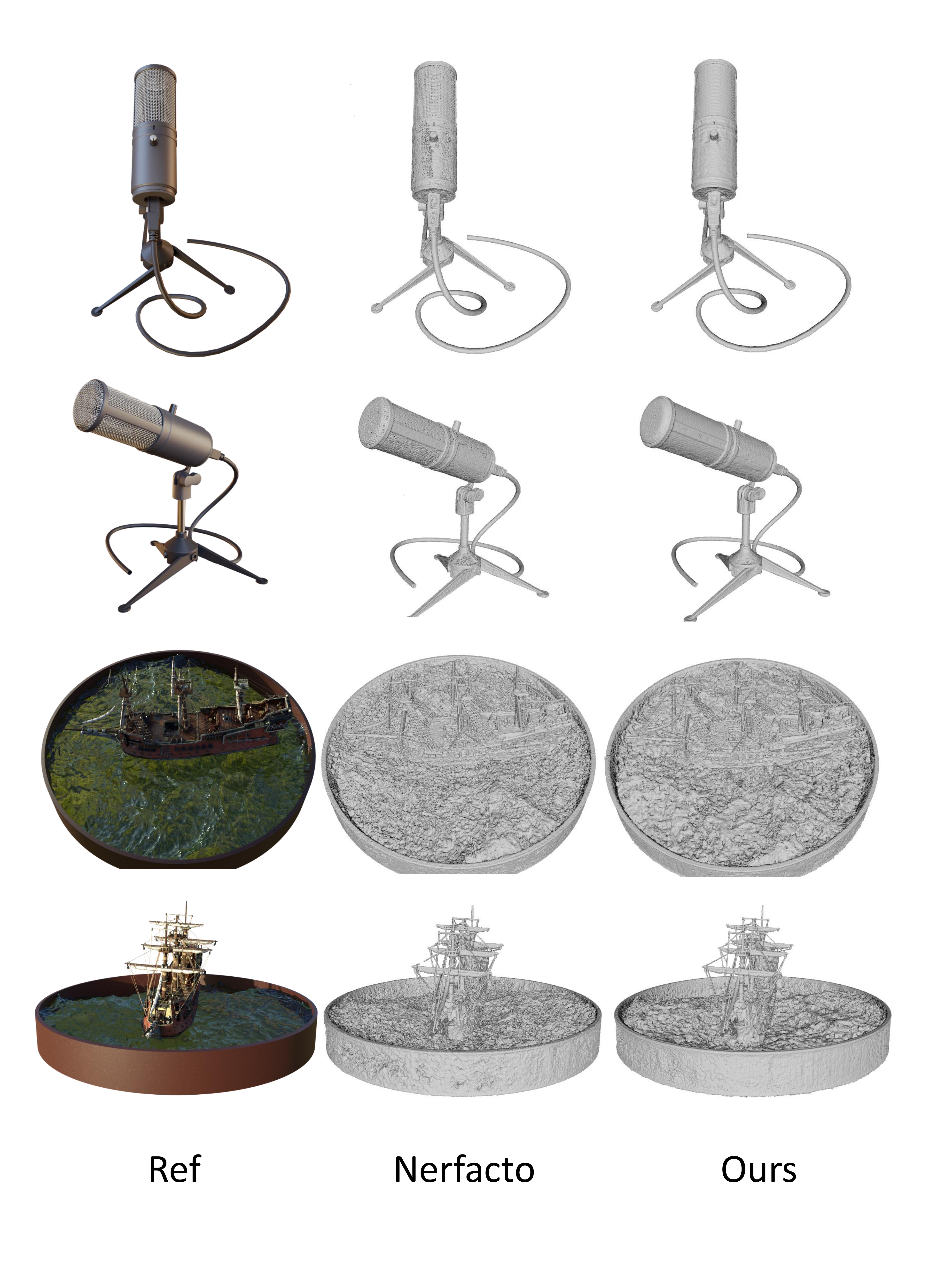}
    \caption{Reconstruction results on the Blender dataset~\cite{mildenhall2020nerf} (Mic, Ship).}
    \label{fig:supp_4}
\end{figure}

\begin{figure*}[ht]
    \centering
    \includegraphics[width = \linewidth]{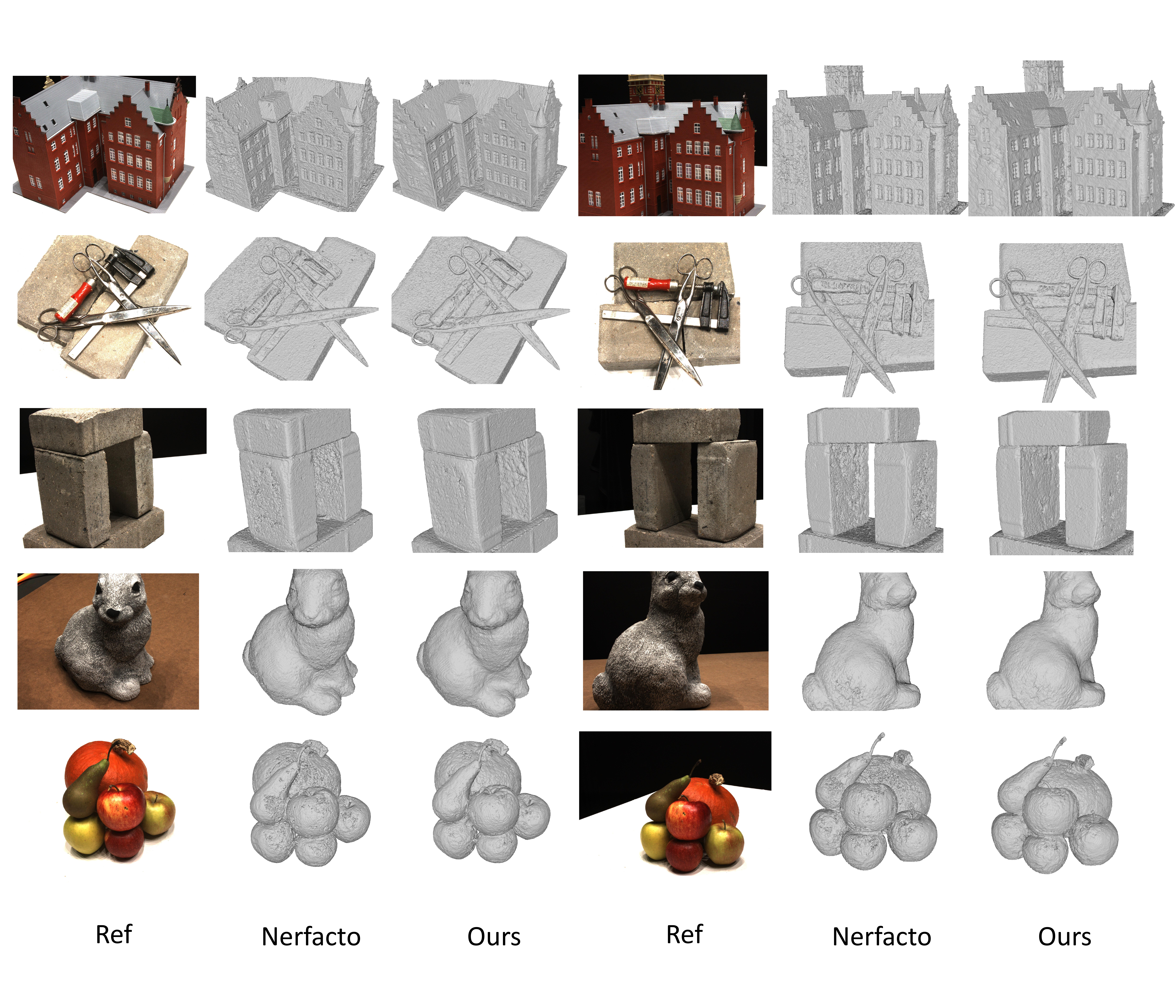}
    \caption{Reconstruction results on the DTU dataset~\cite{jensen2014large}. From top to bottom: Scan 24; Scan 37; Scan 40; Scan 55; Scan 63.}
    \label{fig:supp_dtu_1}
\end{figure*}

\begin{figure*}[ht]
    \centering
    \includegraphics[width = \linewidth]{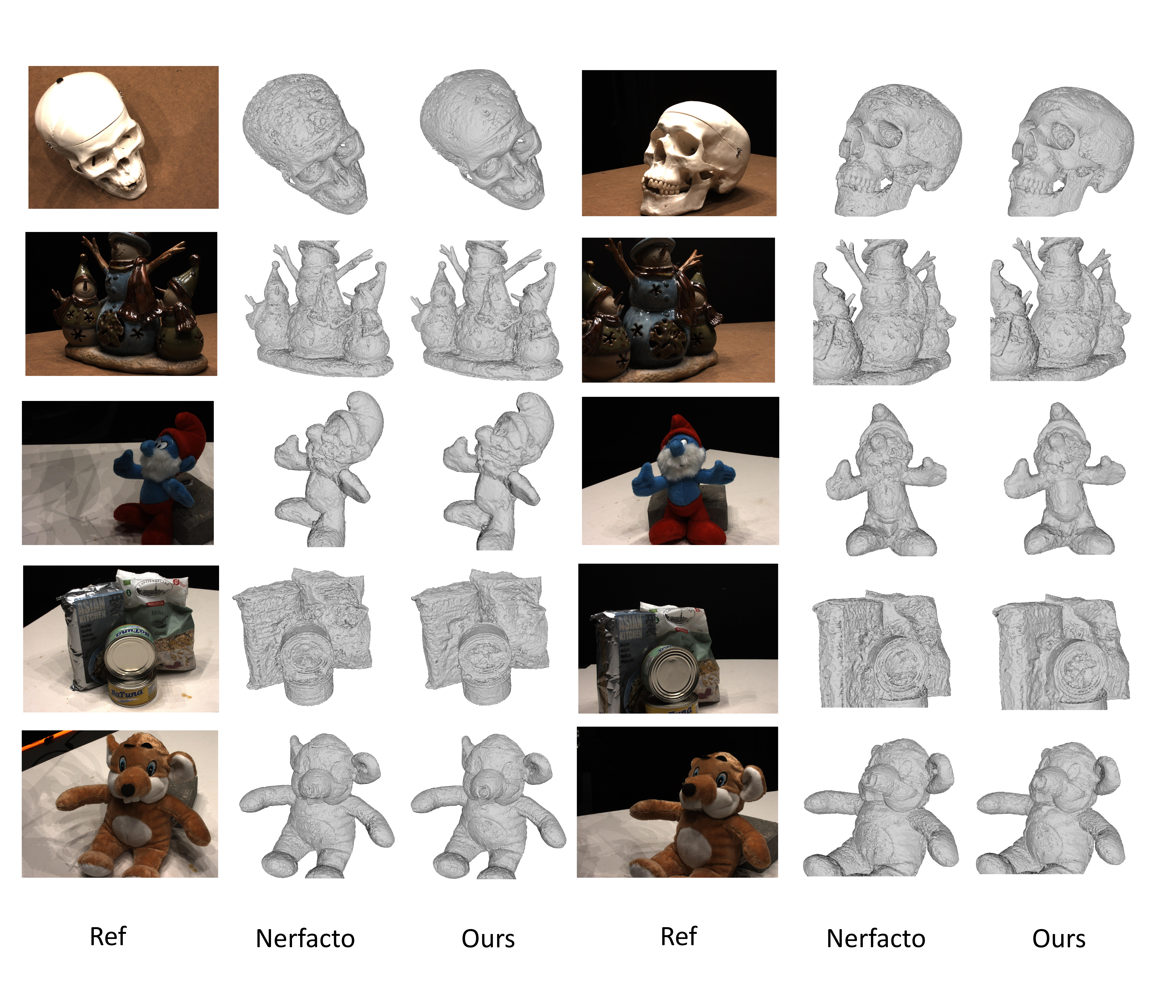}
    \caption{Reconstruction results on the DTU dataset~\cite{jensen2014large}. From top to bottom: Scan 65; Scan 69; Scan 83; Scan 97; Scan 105.}
    \label{fig:supp_dtu_2}
\end{figure*}

\begin{figure*}[ht]
    \centering
    \includegraphics[width = \linewidth]{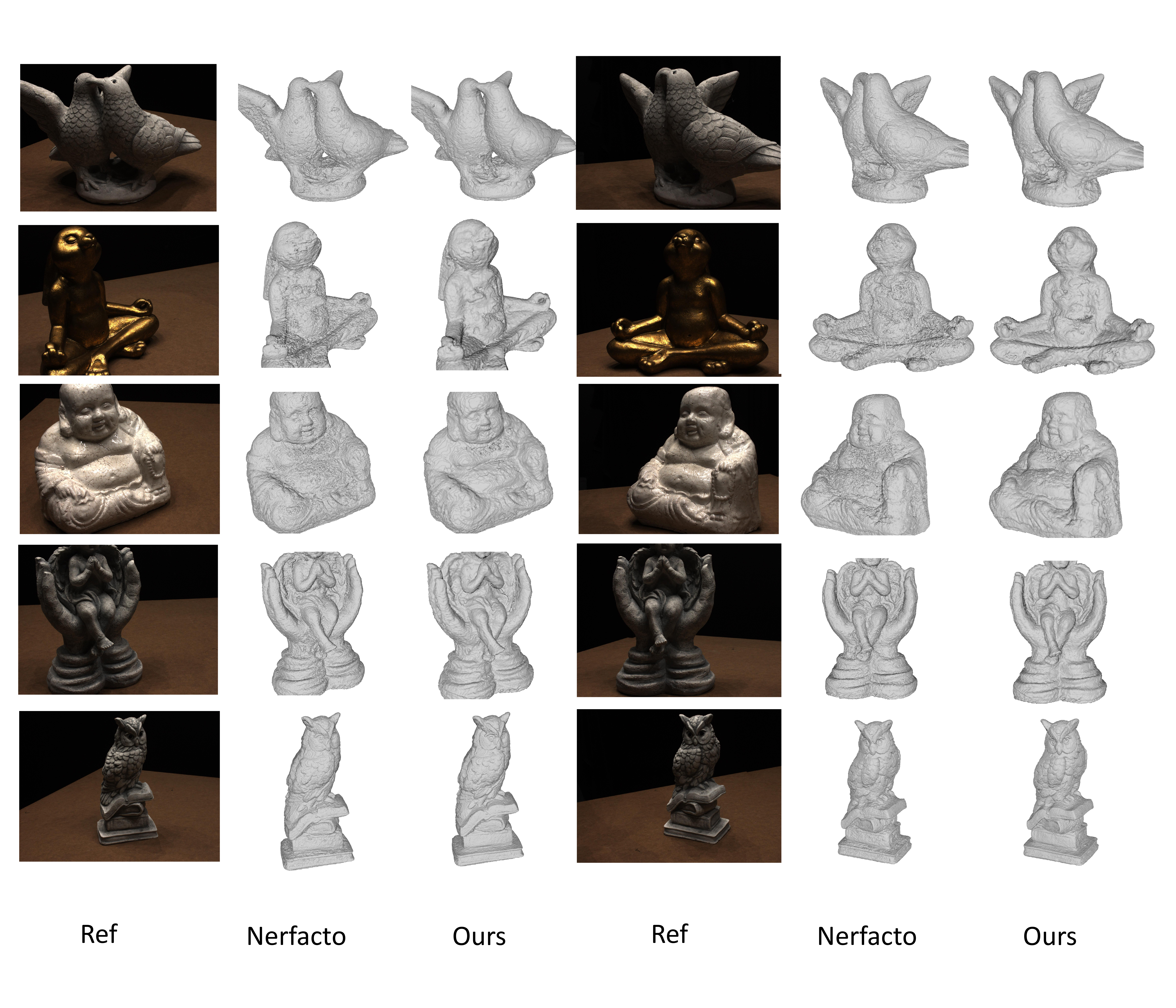}
    \caption{Reconstruction results on the DTU dataset~\cite{jensen2014large}. From top to bottom: Scan 106; Scan 110; Scan 114; Scan 118; Scan 122.}
    \label{fig:supp_dtu_3}
\end{figure*}

Figures (\ref{fig:supp_1}~\ref{fig:supp_2}~\ref{fig:supp_3}~\ref{fig:supp_4}) show all visual comparisons between Nerfacto~\cite{tancik2023nerfstudio} and our method on the Blender dataset~\cite{mildenhall2020nerf}. Figures (\ref{fig:supp_dtu_1}~\ref{fig:supp_dtu_2}~\ref{fig:supp_dtu_3}) present the results of Nerfacto and our approach on the DTU dataset\cite{jensen2014large}. The comprehensive visual comparisons show that our method achieves more accurate geometries compared to the Nerfacto baseline.

\subsection{Reconstruction Results on Transparent Objects}
\begin{figure}[ht]
    \centering
    \includegraphics[width = \linewidth]{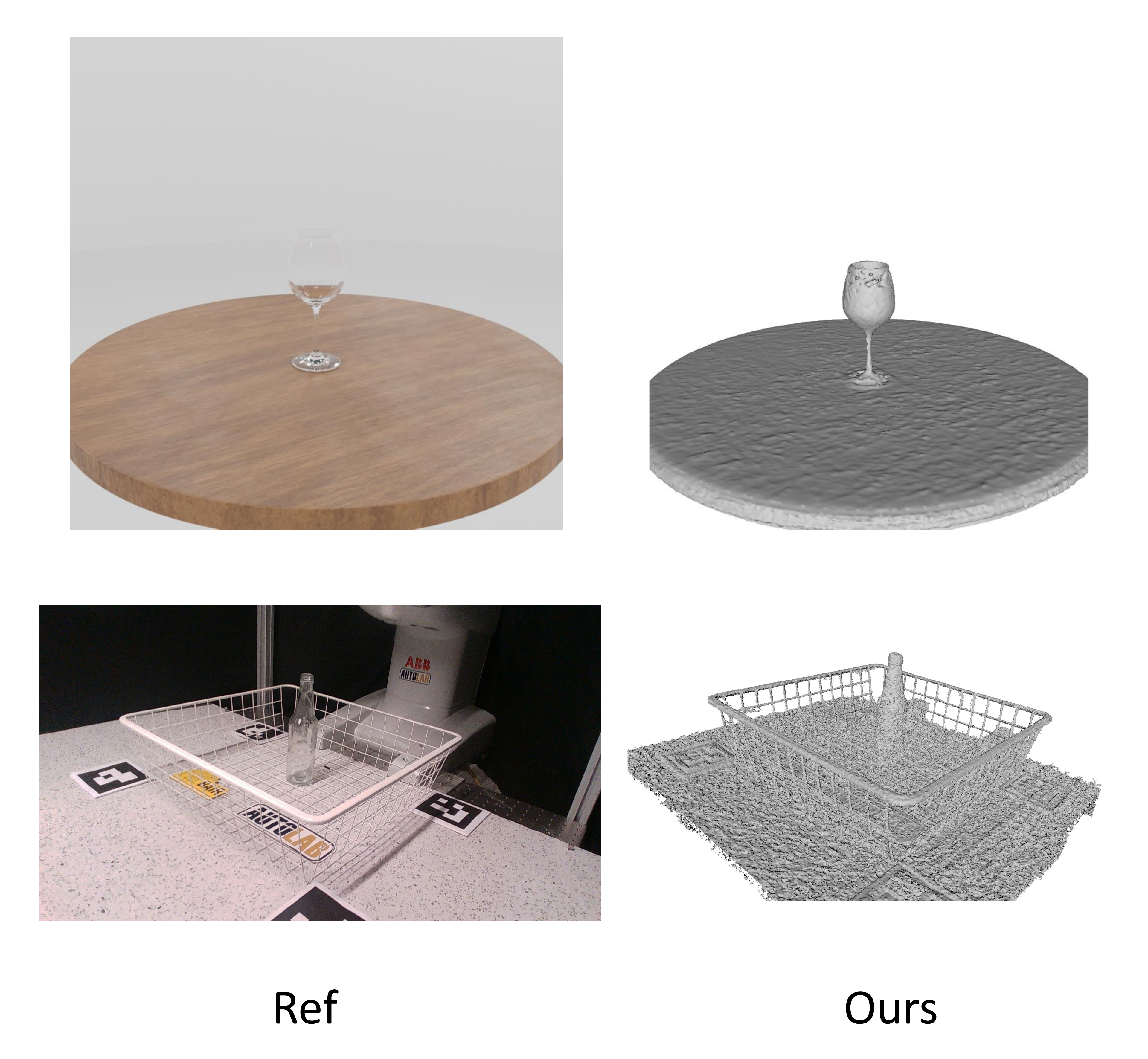}
    \caption{Reconstruction results on the DexNeRF dataset~\cite{IchnowskiAvigal2021DexNeRF}. The first row is a synthetic scene. The second row is a real-world case.}
    \label{fig:dex}
\end{figure}

To further exploit the practicality of our method, we implement our method based on Instant-NGP~\cite{mueller2022instant} and conduct an experiment on 2 scenes of the transparent DexNeRF~\cite{IchnowskiAvigal2021DexNeRF} datasets, shown in Fig.~\ref{fig:dex}

\end{document}